\documentclass[conference]{IEEEtran}
\IEEEoverridecommandlockouts
\usepackage{cite}
\usepackage{amsmath,amssymb,amsfonts}
\usepackage{algorithmic}
\usepackage{graphicx}
\usepackage{textcomp}
\usepackage{xcolor}
\usepackage{hyperref}
\usepackage{booktabs}
\usepackage{multirow}

\hypersetup{
    colorlinks=true,
    linkcolor=blue,   
    citecolor=blue,   
    urlcolor=red      
}
\usepackage[acronym,toc]{glossaries}

\glsdisablehyper

\newacronym{NDS}{NDS}{nuScenes Detection Score}
\newacronym{bev}{BEV}{Bird's Eye View}
\newacronym{mAP}{mAP}{mean Average Precision}
\newacronym{lidar}{LiDAR}{Light Detection and Ranging}
\newacronym{fps}{FPS}{Feature Pyramid Network}
\glsaddall
\def\BibTeX{{\rm B\kern-.05em{\sc i\kern-.025em b}\kern-.08em
    T\kern-.1667em\lower.7ex\hbox{E}\kern-.125emX}}

\begin{document}

\title{Evaluating the Impact of Weather-Induced Sensor Occlusion on BEVFusion for 3D Object Detection\\
\thanks{{\\$^{1}$Dept. of Electronic and Computer Engineering and the Data Driven Computer Engineering Research Centre, University of Limerick, Castletroy, Co. Limerick V94 T9PX, Ireland.} \\ 
$^{2}$Valeo Vision Systems, Tuam, Co. Galway H54 Y276, Ireland. \\
$^{3}$Queen Mary University of London, Mile End Road, London E1 4NS, UK.\\
Corresponding Author: Mr. Sanjay Kumar (\href{mailto:kumar.sanjay@ul.ie}{kumar.sanjay@ul.ie}).

}
}



\author{Sanjay Kumar$^{1}$, Tim Brophy$^{1}$, Eoin Martino Grua$^{1}$, Ganesh Sistu$^{1,2}$, Valentina Donzella$^{3}$, Ciarán Eising$^{1}$}

\maketitle

\begin{abstract}

Accurate 3D object detection is essential for automated vehicles to navigate safely in complex real-world environments. \gls{bev} representations, which project multi-sensor data into a top-down spatial format, have emerged as a powerful approach for robust perception. Although BEV-based fusion architectures have demonstrated strong performance through multimodal integration, the effects of sensor occlusions, caused by environmental conditions such as fog, haze, or physical obstructions, on 3D detection accuracy remain underexplored. In this work, we investigate the impact of occlusions on both camera and \gls{lidar} outputs using the BEVFusion architecture, evaluated on the nuScenes dataset. Detection performance is measured using \gls{mAP} and the \gls{NDS}. Our results show that moderate camera occlusions lead to a 41.3\% drop in mAP (from 35.6\% to 20.9\%) when detection is based only on the camera. On the other hand, \gls{lidar} sharply drops in performance only under heavy occlusion, with mAP falling by 47.3\% (from 64.7\% to 34.1\%), with a severe impact on long-range detection. In fused settings, the effect depends on which sensor is occluded: occluding the camera leads to a minor 4.1\% drop (from 68.5\% to 65.7\%), while occluding \gls{lidar} results in a larger 26.8\% drop (to 50.1\%), revealing the model’s stronger reliance on \gls{lidar} for the task of 3D object detection. Our results highlight the need for future research into occlusion-aware evaluation methods and improved sensor fusion techniques that can maintain detection accuracy in the presence of partial sensor failure or degradation due to adverse environmental conditions.

\end{abstract}


\section{Introduction}

Modern automated vehicles rely on multiple sensors for accurate 3D perception \cite{bib1}. Cameras provide rich appearance cues, such as colour and texture, but lack direct depth information, making it difficult to estimate distances accurately \cite{bib2}. In contrast, \gls{lidar} offers precise geometric measurements and depth data, but produces sparse point clouds that lack the high-resolution detail and semantic richness of camera images \cite{bib3}. Fusing these complementary modalities into a \gls{bev} representation has emerged as an effective strategy for robust 3D object detection \cite{bib4}. Despite progress in BEV-based sensor fusion architectures, most models are still evaluated under ideal, occlusion-free conditions \cite{bib5}, \cite{bib6}. However, in real-world deployments, sensor outputs are frequently degraded due to environmental factors. Cameras may be partially occluded by dirt, raindrops, or smudges on the lens, while \gls{lidar} signals can be disrupted by adverse weather conditions such as fog or heavy rain, which reduce the number of returned points and the perception range \cite{bib7}. These occlusions can significantly degrade detection performance and pose safety risks if not addressed properly. 
\par Previous work has only partially explored this issue. Xie et al. \cite{bib8} investigate camera occlusions but do not consider \gls{lidar} degradation for 3D object detection. Brophy et al. \cite{bib9} analyse the impact of rain on camera-based detection but do not examine the effect of multimodal fusion or occlusion in the \gls{bev}-based domain. The effect of sensor-specific occlusions remains underexplored, particularly in the context of \gls{bev} fusion models. In this work, we systematically investigate the effect of occlusions on both camera and \gls{lidar} sensors using the BEVFusion \cite{bib10} architecture and the nuScenes dataset \cite{bib11}. Camera occlusions are simulated using soiling masks from the Woodscape dataset \cite{bib12} to replicate real-world lens obstructions. For \gls{lidar}, we simulate occlusion by randomly dropping a proportion of the point cloud, mimicking the degradation caused by adverse weather conditions such as fog or rain. We evaluate detection performance using standard metrics, including Mean Average Precision (mAP) and nuScenes Detection Score (NDS), and analyse how different levels of occlusion affect each sensor individually.

The main contributions of this paper are:
\begin{enumerate}
\item We simulate occlusions on both the camera (using the Woodscape soiled dataset) and \gls{lidar} (via random point cloud dropout).
\item We evaluate the impact of these occlusions on 3D object detection using the BEVFusion architecture and the nuScenes dataset.
\item We analyse modality-specific and fused performance under occlusion, highlighting the limitations of current fusion models and motivating the need for occlusion-aware fusion strategies.
\end{enumerate}

The remainder of the paper is structured as follows: Section 2 reviews related work. Section 3 describes the nuScenes dataset and explains how the camera and \gls{lidar} occlusions are applied. Section 4 outlines the methodology, including the BEVFusion model architecture and experimental setup. Section 5 presents the results of our experiments. Section 6 discusses key findings and future directions.

\section{Related Work}

3D object detection has become a critical task in automated driving, with methods generally categorised by their sensor inputs: camera-only, \gls{lidar}-only, and multi-modal fusion of camera and \gls{lidar} data.

\subsection{Camera-Only Methods}

Recent advances have shown that 3D detection is possible using only image data. Li et al. \cite{bib13} and Xie et al. \cite{bib14} are two prominent researchers who developed approaches that construct a \gls{bev} representation from multi-camera images to facilitate detection. Li et al. \cite{bib13} introduce a spatiotemporal transformer that aggregates camera views into a unified \gls{bev} space. In contrast, Xie et al. \cite{bib14} jointly perform 3D detection and map segmentation by projecting multi-view image features into a shared voxel-based \gls{bev} grid. These methods avoid explicit depth supervision but still achieve competitive accuracy using camera-only inputs. However, camera-only methods struggle with accurate depth estimation, especially in low-light, distant, or occluded scenes. Without direct depth sensing, their 3D localisation is less reliable in challenging driving conditions.

\subsection{\gls{lidar}-Only Methods}

\gls{lidar}-based approaches benefit from precise geometric information. Lang et al. \cite{bib15} encodes point cloud data into a pseudo-image using vertical columns (pillars), allowing fast inference via 2D convolutions in \gls{bev} space. Yin et al. \cite{bib16} further improve performance by adopting a centre-based detection framework, where object centres are predicted directly from \gls{bev} features, simplifying the regression pipeline and enhancing both accuracy and speed. However, \gls{lidar}-only methods lack the rich semantic and texture information available in camera images, making it difficult to distinguish between visually similar objects and accurately interpret scene context, especially for small or distant objects.

\subsection{Camera–\gls{lidar} Fusion Methods}
To leverage the complementary strengths of \gls{lidar} geometry and camera semantics, various fusion-based approaches have been proposed. Chen et al. \cite{bib17} introduce learnable spatial offsets to align pixel-level image features with 3D \gls{lidar} representations, enhancing the fusion process in \gls{bev}. Xu et al. \cite{bib18} build on this by selectively injecting camera semantics into point cloud features via adaptive attention, improving performance, particularly for small or distant objects. Bai et al. \cite{bib19} present a more robust and flexible fusion strategy: it first generates initial 3D bounding boxes using \gls{lidar} \gls{bev} features, then refines predictions through a transformer-based attention mechanism that softly integrates relevant camera information, making it resilient to sensor misalignment and poor lighting conditions. In contrast, Chen et al. \cite{bib20} avoid the \gls{bev} representation altogether by using a transformer to fuse camera and \gls{lidar} features directly into 3D space, enabling the end-to-end prediction of bounding boxes from multimodal data. Collectively, these methods demonstrate that tightly coupled and learnable fusion mechanisms significantly boost 3D detection performance.

\subsection{Robustness under Adverse Weather and Occlusion}
While recent fusion methods have advanced detection accuracy on clean benchmarks, robustness under adverse weather and occlusion has received comparatively less attention. Xie et al. \cite{bib8} introduced a benchmark to evaluate camera-based \gls{bev} models under natural corruptions such as fog, snow, low light, and motion blur, showing that detection performance drops sharply and varies across architectures. Kumar et al. \cite{bib21} studied occlusions caused by dirt, raindrops, and fog in camera inputs, demonstrating significant degradation in \gls{bev} perception and highlighting the benefit of incorporating LiDAR and radar for robustness. More recently, Wang et al. \cite{bib22} proposed MSAFusion, a reinforcement learning–based adaptive \gls{bev} fusion framework that dynamically adjusts sensor weights and uses random modality dropout to improve resilience under adverse weather, validated on nuScenes and Radiate datasets. Collectively, these works emphasise that adverse conditions substantially affect perception, yet systematic robustness evaluations of established \gls{bev} fusion models remain limited.

\begin{figure*}[t]
  \centering
  \includegraphics[width=\textwidth]{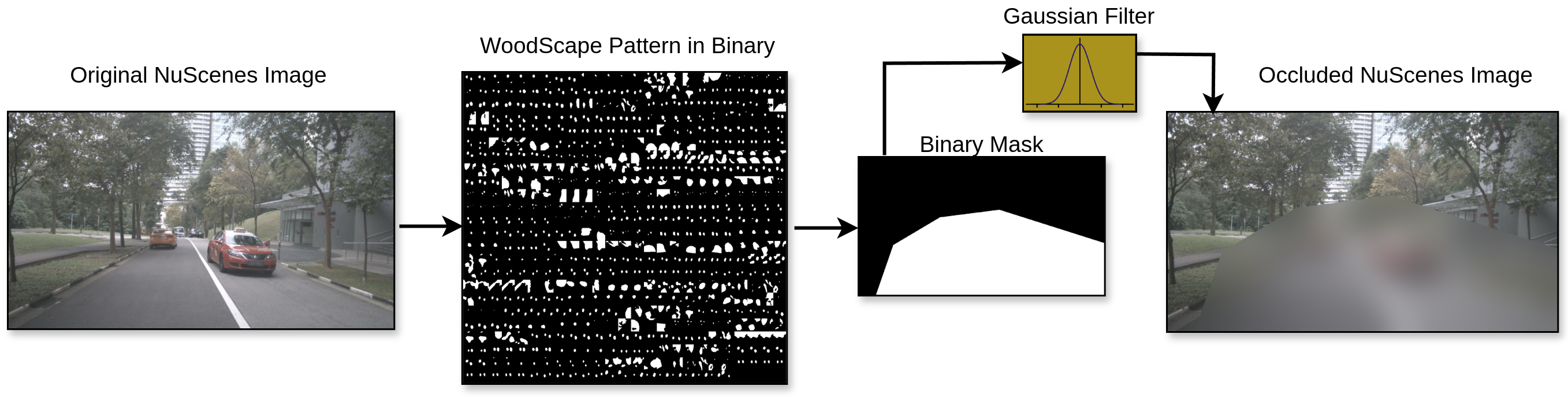}
  \vspace{-7mm}
  \caption{From left to right: the original nuScenes image, followed by multiple binary occlusion patterns from the WoodScape soiling dataset. A randomly selected binary mask is smoothed using a Gaussian filter and applied to the original image to produce an occluded nuScenes image.}
  \label{Figure:cam_occ}
  \vspace{-5mm}
\end{figure*}

\subsection{Research Gap}
Although extensive progress has been made in camera-only, \gls{lidar}-only, and fusion methods, these approaches are typically trained and evaluated on clean, unoccluded data. Prior robustness studies address camera-only degradations, but they do not systematically examine existing \gls{bev} fusion models under controlled sensor occlusion. To fill this gap, we apply structured occlusions to both camera and LiDAR inputs in nuScenes and evaluate the robustness of BEVFusion for 3D object detection in \gls{bev} space.

\section{Dataset}
\label{dataset}
We use the nuScenes dataset \cite{bib11}, a large-scale benchmark for automated driving comprising 1,000 urban driving scenes from Boston and Singapore. Each 20-second scene includes synchronised data from six cameras, five radars, and a 32-beam \gls{lidar}, sampled at 2 Hz, with 3D annotations for multiple object classes. The dataset provides 28,130 training and 6,019 validation samples, along with detailed maps, ego-vehicle poses, and calibrated sensor parameters, making it ideal for \gls{bev}-based multi-sensor perception. We chose nuScenes over other datasets, such as KITTI \cite{bib23} and Waymo \cite{bib24}, due to its $360^\circ$ multi-sensor setup (including radar) and its diverse urban scenarios with varying weather and lighting conditions. In comparison, KITTI lacks radar and has only front-facing sensors in clear conditions, while Waymo also lacks radar and offers less variation in environment and lighting.
While nuScenes offers the advantage of 360$^{\circ}$ calibrated sensor coverage, real deployments may involve partial fields of view or non-overlapping camera–\gls{lidar} angles. This limitation is important, as fusion performance can degrade in the presence of blind spots or misalignment. Addressing such cases is part of our ongoing work, where we study occlusions applied to specific regions of the \gls{lidar} field of view.

To investigate the impact of sensor occlusion on 3D object detection in \gls{bev}, we simulate occlusions on both the camera and \gls{lidar} modalities. For camera occlusion, we apply binary soiling masks derived from the Woodscape soiled dataset \cite{bib12} to simulate real-world visual obstructions such as fog on the lens. For \gls{lidar}, we simulate occlusion by randomly dropping a percentage of point cloud data, mimicking degradation caused by reduced visibility. These controlled degradations allow us to analyse the sensitivity of \gls{bev}-based detection architectures to real-world sensor challenges. \\

\subsection{Simulated Camera Degradation}

To simulate camera occlusion in the nuScenes dataset, as illustrated in Figure~\ref{Figure:cam_occ}, we apply synthetic soiling patterns that resemble common visual degradations such as fog on the lens. These occlusion masks are derived from the Woodscape Soiled dataset, which provides a diverse set of contamination patterns observed in real driving conditions.

We begin by generating binary masks from these patterns to identify the regions of each camera image affected by occlusion. To simulate the visual distortion caused by lens contamination, we apply a Gaussian filter specifically to the masked areas. This replicates how fog blurs the visibility of the affected image regions while keeping the remaining image content intact.

Mathematically, the blurred occluded region \( I_{\text{occluded}} \) is computed as:
\[
I_{\text{occluded}} = G_{\sigma} * (I \odot M)
\]
where \( I \) is the original image, \( M \) is the binary occlusion mask (1 for occluded pixels, 0 elsewhere), \( \odot \) denotes element-wise multiplication, \( G_{\sigma} \) is a Gaussian kernel with standard deviation \( \sigma \), and \( * \) represents the convolution operation.

The final occluded image \( I' \) is obtained by blending the blurred occluded regions with the clean parts of the image:
\[
I' = I \odot (1 - M) + I_{\text{occluded}}
\]
This method produces localised image degradation, preserving the global structure of the scene while impairing only the occluded areas. It enables a controlled study of how camera occlusion impacts \gls{bev}-based 3D object detection performance.

\subsection{Simulated \gls{lidar} Degradation}

\begin{figure*}[t]
  \centering
  \includegraphics[width=\textwidth]{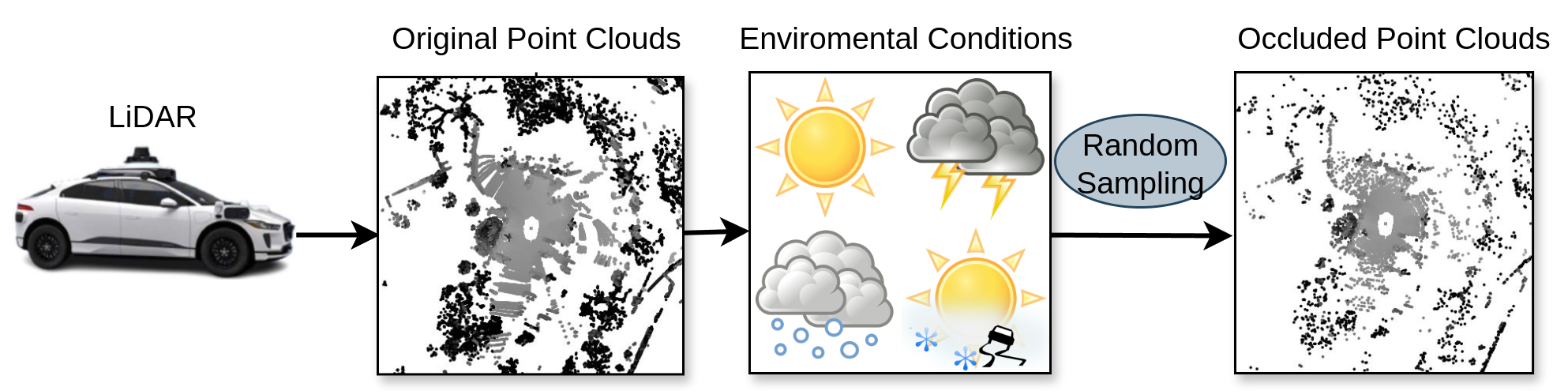}
  \vspace{-4mm}
  \caption{From left to right: original nuScenes \gls{lidar} point clouds, simulated environmental conditions (e.g., rain, fog, snow), a random sampling step that removes some \gls{lidar} points, and the resulting occluded point clouds.}
  \label{Figure:lid_occ}
  \vspace{-5mm}
\end{figure*}

To investigate the effects of \gls{lidar} occlusion on \gls{bev}-based 3D object detection, we simulate point cloud degradation in the nuScenes dataset by randomly removing a portion of \gls{lidar} points from each frame, as shown in Figure~\ref{Figure:lid_occ}. This random dropout mimics real-world scenarios in which \gls{lidar} signals are partially lost due to environmental factors such as fog, storm, and rain. Given an input point cloud with \( N \) points, a fixed dropout ratio \( r \in [0, 1] \) is used to control the severity of degradation. The number of retained points is computed as:
\[
N_{\text{retained}} = N \times (1 - r)
\]
This method allows for controlled occlusion while avoiding spatial bias, as points are dropped uniformly at random across the scene. By varying \( r \), we simulate different levels of occlusion severity and analyse the robustness of \gls{bev}-based detection models under degraded \gls{lidar} input conditions. This approach aligns with Chan et al. \cite{bib25}, who simulate LiDAR occlusion by removing beams to model environmental noise factors like rain and partial lens blockage. We use simulated occlusions since nuScenes lacks sufficient severe fog, rain, or soiling cases for controlled analysis. Soiling masks and point dropout approximate these real-world degradations while allowing systematic and repeatable evaluation.

\section{METHODOLOGY}
In this study, we evaluate the robustness of the BEVFusion \cite{bib10} architecture under sensor occlusions to understand how degradation in camera and \gls{lidar} inputs affects 3D object detection performance. We chose BEVFusion as it processes camera and \gls{lidar} modalities independently, making it suitable for isolating occlusion effects, while remaining computationally efficient compared to newer transformer-based fusion models. BEVFusion is a state-of-the-art multi-sensor fusion framework that projects features from multiple modalities (camera and \gls{lidar}) into a unified \gls{bev} representation. This allows effective fusion for the downstream perception task, i.e., 3D object detection.

\subsection{Architecture Overview}

\begin{figure*}[h!]
\centering
\includegraphics[width=\textwidth]{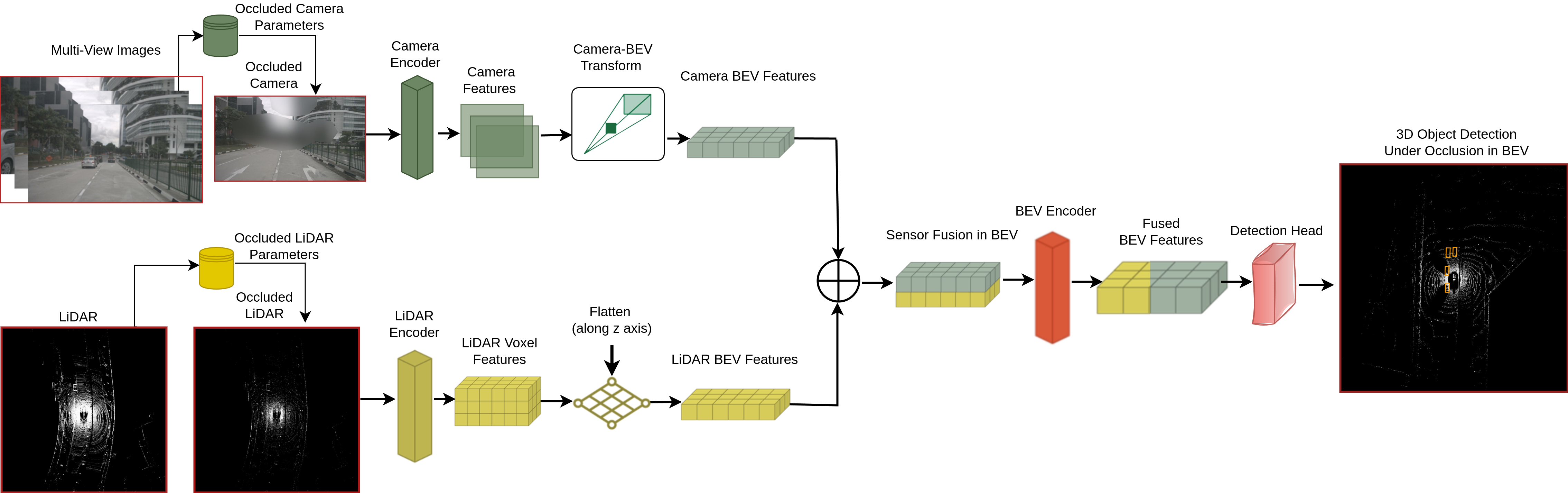}
\vspace{-4mm}
\caption{Overview of the BEVFusion architecture used in our study. From left to right: multi-view camera and \gls{lidar} inputs (including occluded variants) are encoded and transformed into \gls{bev} space, followed by sensor fusion, \gls{bev} encoding, and 3D object detection under occlusion.} 
\label{fig:ModifiedBEVFusion}
\vspace{-5mm}
\end{figure*}

BEVFusion transforms multi-view camera images and \gls{lidar} point clouds into a shared \gls{bev} space through steps involving modality-specific encoders, view transformation to \gls{bev} space, \gls{bev} feature fusion, and task-specific prediction heads. As demonstrated in Figure~\ref{fig:ModifiedBEVFusion}, we describe each component in detail below:


\subsubsection{Camera Encoder} Each input camera image is processed independently by a convolutional backbone, Swin-T \cite{bib26}, to extract 2D feature maps rich in semantic information. These features are further refined and merged across scales using a \gls{fps} \cite{bib27}, resulting in a unified multi-scale representation. The final output is a downsampled feature map of shape:
\[
F^{\text{cam}} \in \mathbb{R}^{N \times H \times W \times C}
\]
where \( N \) is the number of camera views, \( H \times W \) is the spatial resolution of the feature map, and \( C \) is the number of feature channels.

\subsubsection{\gls{lidar} Encoder} Raw \gls{lidar} point clouds are voxelized and passed through a 3D convolutional backbone, VoxelNet \cite{bib28}, to extract spatial features. The extracted \gls{lidar} features are represented as:
\[
F^{\text{lidar}}_{\text{3D}} \in \mathbb{R}^{X \times Y \times Z \times C}
\]
where \( X, Y, Z \) are the voxel grid dimensions and \( C \) is the number of feature channels.


\subsubsection{Camera-to-\gls{bev} View Transformation} The camera feature maps, originally in perspective view, are projected into 3D space using depth-aware lifting \cite{bib27}. Specifically, each pixel feature is lifted along its camera ray into a set of \( D \) discrete depth bins. A depth distribution is predicted per pixel, and camera features are projected into 3D space according to:
\[
F^{\text{cam}}_{\text{3D}}(x, y, z) = \sum_{d=1}^{D} p_d \cdot f_{(u,v,d)}
\]
where \( p_d \) is the predicted probability for depth bin \( d \), and \( f_{(u,v,d)} \) is the lifted camera feature at pixel \( (u,v) \) and depth \( d \). This operation generates a volumetric 3D feature representation from 2D image features.

\subsubsection{\gls{bev} Pooling (Camera)} The lifted 3D camera features are collapsed along the vertical (\( z \)) axis to form a 2D \gls{bev} representation. BEVFusion uses sum pooling to aggregate features within each vertical column:
\[
F^{\text{cam}}_{\text{BEV}}(x, y) = \sum_{z} F^{\text{cam}}_{\text{3D}}(x, y, z)
\]
This pooling strategy preserves the total feature magnitude within each grid cell and supports a dense accumulation of semantic features.

\subsubsection{\gls{lidar}-to-\gls{bev} Projection} Similarly, the 3D \gls{lidar} features are projected into \gls{bev} by summing over the vertical axis. In practice, BEVFusion implements this via an efficient prefix sum over the voxelized features:
\[
F^{\text{lidar}}_{\text{BEV}}(x, y) = \sum_{z} F^{\text{lidar}}_{\text{3D}}(x, y, z)
\]
This transformation aligns the \gls{lidar} features to the \gls{bev} plane and prepares both modalities for spatially consistent fusion in the next stage.

\subsubsection{\gls{bev} Feature Fusion}
The \gls{bev} features from both modalities are concatenated to form a unified \gls{bev} representation:
\[
F^{\text{fused}}_{\text{BEV}} = \mathcal{E}_{\text{BEV}} \left( F^{\text{cam}}_{\text{BEV}} \oplus F^{\text{lidar}}_{\text{BEV}} \right)
\]
where \( \oplus \) denotes feature concatenation along the channel dimension, and \( \mathcal{E}_{\text{BEV}} \) is a fully convolutional \gls{bev} encoder. This encoder applies residual convolutional blocks to fuse the concatenated features and to mitigate any spatial misalignment between modalities.

\subsubsection{Task-Specific Prediction Head}
The fused \gls{bev} features are passed to task-specific heads for downstream 3D perception tasks. For 3D object detection, BEVFusion adopts a head structure inspired by \cite{bib15} \cite{bib18}, which includes:

\begin{itemize}
    \item A \textbf{center heatmap head} to predict the 2D location of object centers in the \gls{bev} space.
    \item \textbf{Regression heads} to estimate object attributes such as size, orientation, and velocity.
\end{itemize}

Each head is implemented as a lightweight convolutional block applied to the fused \gls{bev} feature map. These predictions are jointly supervised using standard object detection losses during training. Our setup utilises the pre-trained detection head provided by BEVFusion and evaluates its performance under various sensor occlusion settings without requiring retraining.

\subsubsection{Evaluation Under Sensor Occlusion}

We evaluate the performance of the unmodified BEVFusion architecture on the nuScenes validation set under simulated sensor occlusions. Occlusions are applied to the camera and \gls{lidar} inputs independently during inference, without retraining the model. The occlusion generation process, including Woodscape soiling masks for camera images and random point dropout for \gls{lidar}, is described in detail in Section \ref{dataset}. The goal is to assess the out-of-the-box robustness of BEVFusion to sensor degradation. Performance is measured using mean Average Precision (mAP) and nuScenes Detection Score (NDS), with analysis across different \gls{lidar} occlusion levels.\\

\section{EXPERIMENTAL RESULTS}

\begin{table*}[ht]
\centering
\caption{BEVFusion performance under increasing \gls{lidar} occlusion (0\%–90\%) with clean camera input. Results compare \gls{lidar}-only vs. Camera+\gls{lidar} occlusion in terms of mAP and NDS.}
\vspace{-2mm}
\renewcommand{\arraystretch}{1.1}
\begin{tabular}{@{}c cc cc cc cc cc cc@{}}
\toprule
\textbf{Occluded Sensor} & \multicolumn{2}{c}{\textbf{0\%}} & \multicolumn{2}{c}{\textbf{30\%}} & \multicolumn{2}{c}{\textbf{60\%}} & \multicolumn{2}{c}{\textbf{70\%}} & \multicolumn{2}{c}{\textbf{80\%}} & \multicolumn{2}{c}{\textbf{90\%}} \\
\cmidrule(lr){2-3} \cmidrule(lr){4-5} \cmidrule(lr){6-7} \cmidrule(lr){8-9} \cmidrule(lr){10-11} \cmidrule(lr){12-13}
& mAP & NDS & mAP & NDS & mAP & NDS & mAP & NDS & mAP & NDS & mAP & NDS \\
\midrule
\gls{lidar}              & 64.68 & 69.28 & 62.58 & 67.98 & 56.94 & 64.38 & 54.65 & 62.96 & 48.06 & 58.80 & 34.09 & 49.62 \\
Camera + \gls{lidar}     & 68.52 & 71.38 & 67.26 & 70.41 & 63.95 & 68.16 & 62.90 & 67.40 & 59.08 & 64.59 & 50.09 & 57.94 \\
\bottomrule
Camera is Clean, \gls{lidar} is Occluded
\end{tabular}
\label{lidar_more_experiments}
\vspace{-5mm}
\end{table*}

\subsection{Qualitative Results}

\subsubsection{\gls{lidar} Occlusion with Clean Camera}

We qualitatively evaluate the effect of \gls{lidar} occlusion on 3D object detection performance while maintaining a clean camera input. The visualisation showcases detection results under different sensor configurations, with multi-view camera images on the left and corresponding \gls{bev} outputs on the right combined with \gls{lidar} scan.\\

\begin{figure}[h!]
    \vspace{-3mm}
    \centering
    \includegraphics[width=\columnwidth]{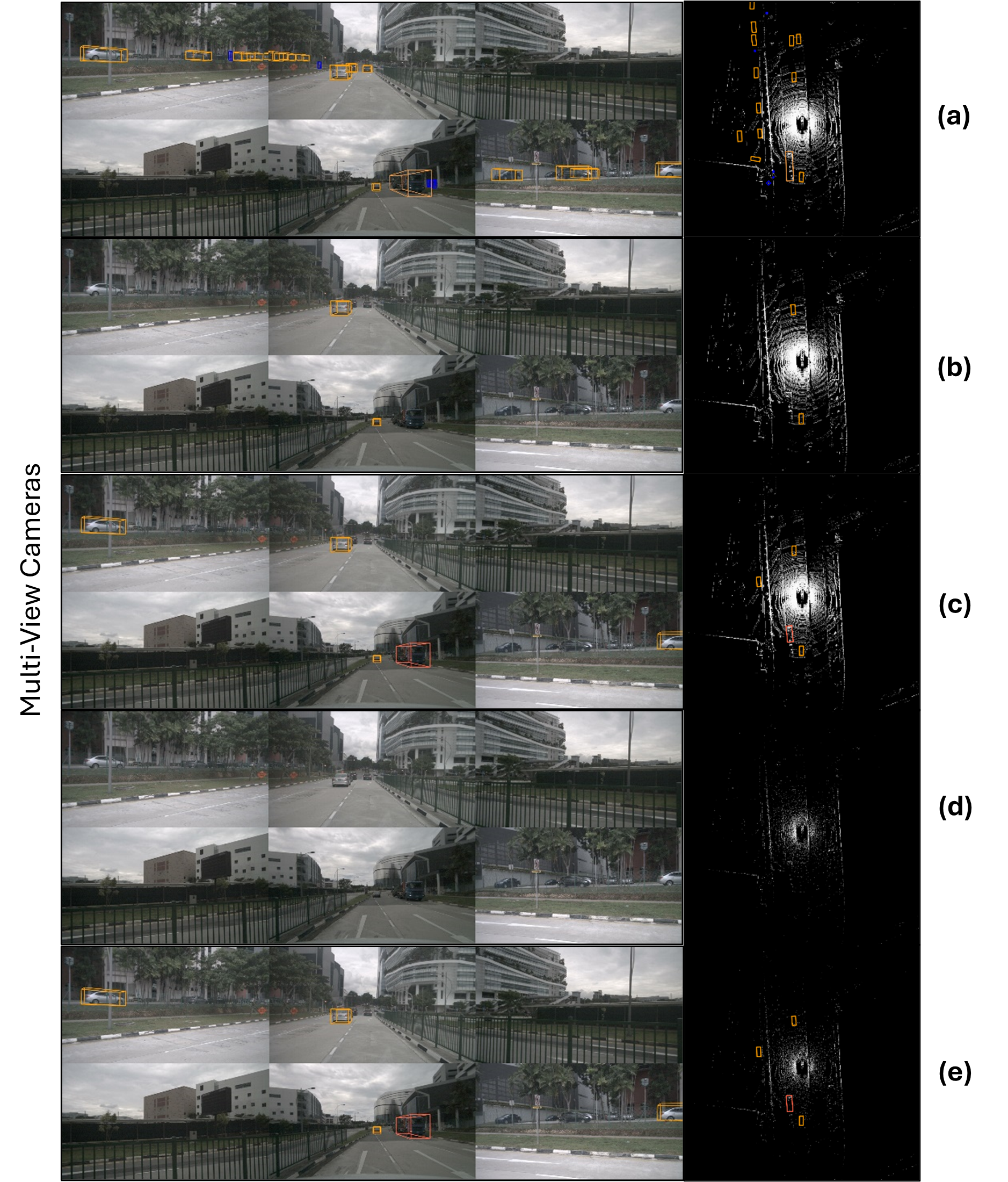}
    \vspace{-3mm}
    \caption{
    Qualitative comparison of BEVFusion predictions under \gls{lidar} occlusion. From top to bottom: (a) Ground truth, (b) Prediction with clean camera input, (c) Prediction with clean \gls{lidar} input, (d) Prediction with occluded \gls{lidar}, and (e) Prediction with clean camera + occluded \gls{lidar}. For each row, the left images show the multi-view camera inputs, with the 3D detection predictions overlaid. The right image displays the input \gls{lidar} scan, with the predictions overlaid in the \gls{bev} space. 
    }
    \label{fig:lidar_occlusion}
    \vspace{-3mm}
\end{figure}

In Figure \ref{fig:lidar_occlusion}, the top row presents the ground truth for reference. When the camera and \gls{lidar} are clean, the model successfully detects and localises objects in the scene with high precision. Following this, with clean camera input, the camera-only sensor (second row) cannot detect several objects due to the lack of depth cues, highlighting the limitations of vision-only perception. The third and fourth rows show predictions using \gls{lidar} input only: the third row uses clean \gls{lidar} data, while the fourth simulates occlusion by dropping 90\% of \gls{lidar} points. Although camera images are included in these two rows for visualisation purposes, they are not used during inference. These views help the reader better understand the layout of the scene and the positions of the objects, even though predictions are made solely from \gls{lidar} data.

With clean \gls{lidar} input alone (third row), the model maintains strong localisation accuracy, benefiting from precise spatial measurements. However, in the following row (fourth row), where \gls{lidar} is progressively occluded by randomly dropping 90\% of the point clouds, almost all objects are missed. As the distance between the object and the \gls{lidar} sensor increases, the perception range decreases, becoming insufficient for reliable detection. Notably, in the fusion configuration (fifth row), where clean camera input is combined with 90\% occluded \gls{lidar}, the model consistently outperforms the \gls{lidar}-only counterpart (d). The preserved visual features help compensate for the loss of geometric structure, allowing the model to detect missed objects in the \gls{lidar}-only (d) configuration under occlusion. This demonstrates the effectiveness of BEVFusion’s multi-modal design, which integrates camera semantics and \gls{lidar} geometry into a unified \gls{bev} representation. Even under severe \gls{lidar} degradation, the fusion framework enables more robust detection by leveraging complementary strengths from both modalities.\\

\subsubsection{Camera Occlusion with Clean \gls{lidar}}

To evaluate the effect of camera degradation, we analyse detection performance when the camera input is occluded while the \gls{lidar} input remains clean. The visualisation shows multi-view camera images on the left and \gls{bev} outputs on the right combined with \gls{lidar} scan, comparing predictions across different sensor setups, i.e., camera and \gls{lidar}.

\begin{figure}[ht]
    \centering
    \includegraphics[width=\columnwidth]{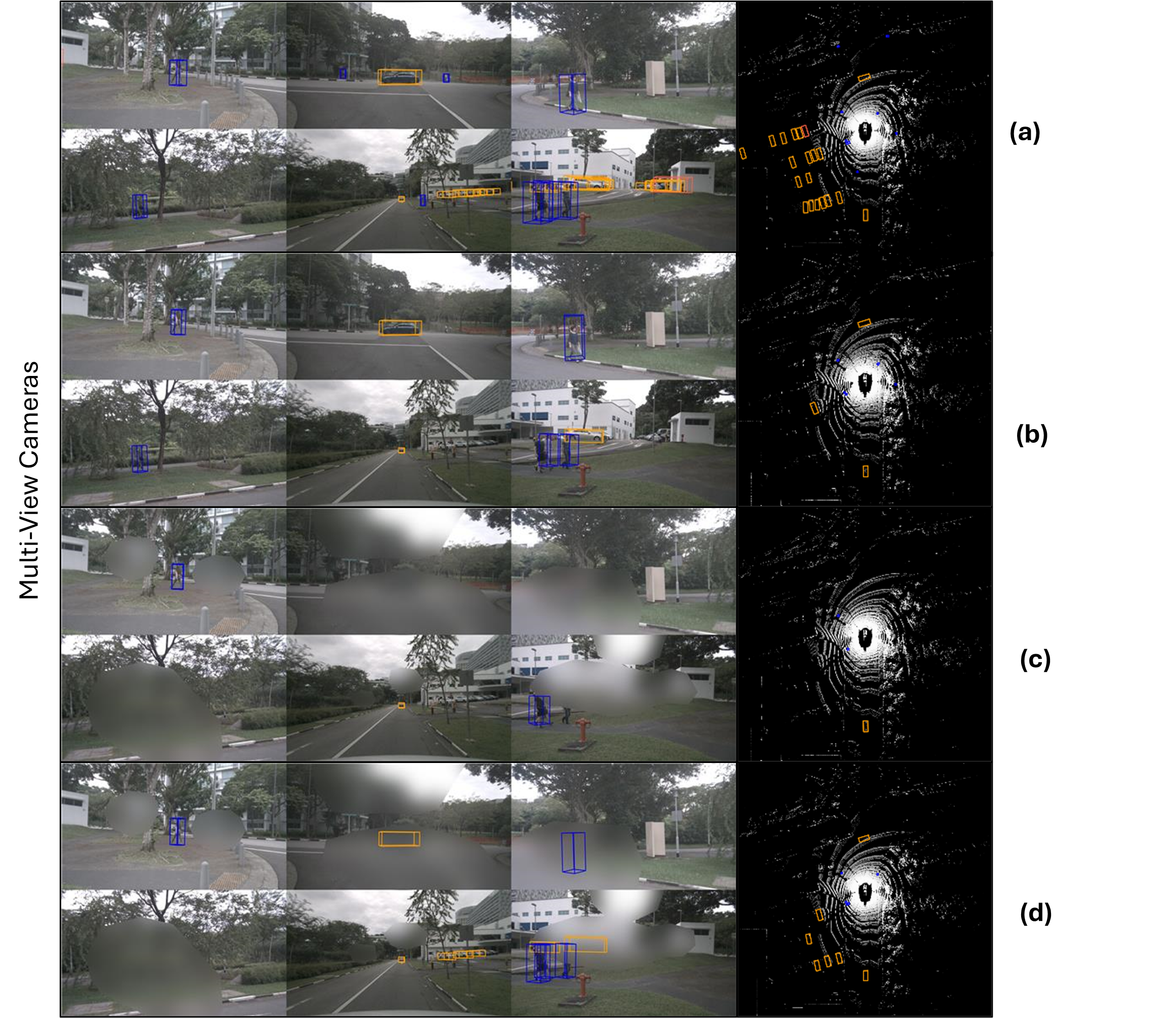}
    \vspace{-3mm}
    \caption{
    Qualitative comparison of BEVFusion predictions under camera occlusion. (a) Ground truth, (b) Prediction with clean camera, (c) Prediction with occluded camera, and (d) Prediction with occluded camera + clean \gls{lidar}. For each row, the left images show the multi-view camera inputs, with the 3D detection predictions overlaid. The right image displays the input \gls{lidar} scan, with the predictions overlaid in the \gls{bev} space.
    }
    \label{fig:camera_occlusion}
    \vspace{-3mm}
\end{figure}

In Figure \ref{fig:camera_occlusion}, the top row shows the ground truth annotations for reference. The second row presents the output when only the camera input is used and kept clean. In this setting, the model can detect several objects using rich semantic features from the visual stream, but it struggles with accurate localisation, particularly for distant or partially occluded objects, due to the lack of depth information. The third row shows results when only the camera is used, and it is occluded. With no \gls{lidar} input, detection performance drops significantly. Many objects are missed or inaccurately localised, especially those farther from the ego vehicle. This highlights the vulnerability of vision-only models under adverse visual conditions. In the following rows, we analyse the fusion configuration where the camera is occluded but the \gls{lidar} input remains clean. Despite visual degradation simulating real-world obstructions, such as fog, the model continues to detect and localise most objects accurately. This is due to the strong spatial information provided by the clean \gls{lidar} sensor, which compensates for the missing or corrupted semantic features from the camera. These results confirm the robustness of the BEVFusion architecture under camera degradation when reliable \gls{lidar} input is available.\\

\subsubsection{Key Observation under \gls{lidar} Occlusion}

\begin{figure}[h!]
    \centering
    \includegraphics[width=\columnwidth]{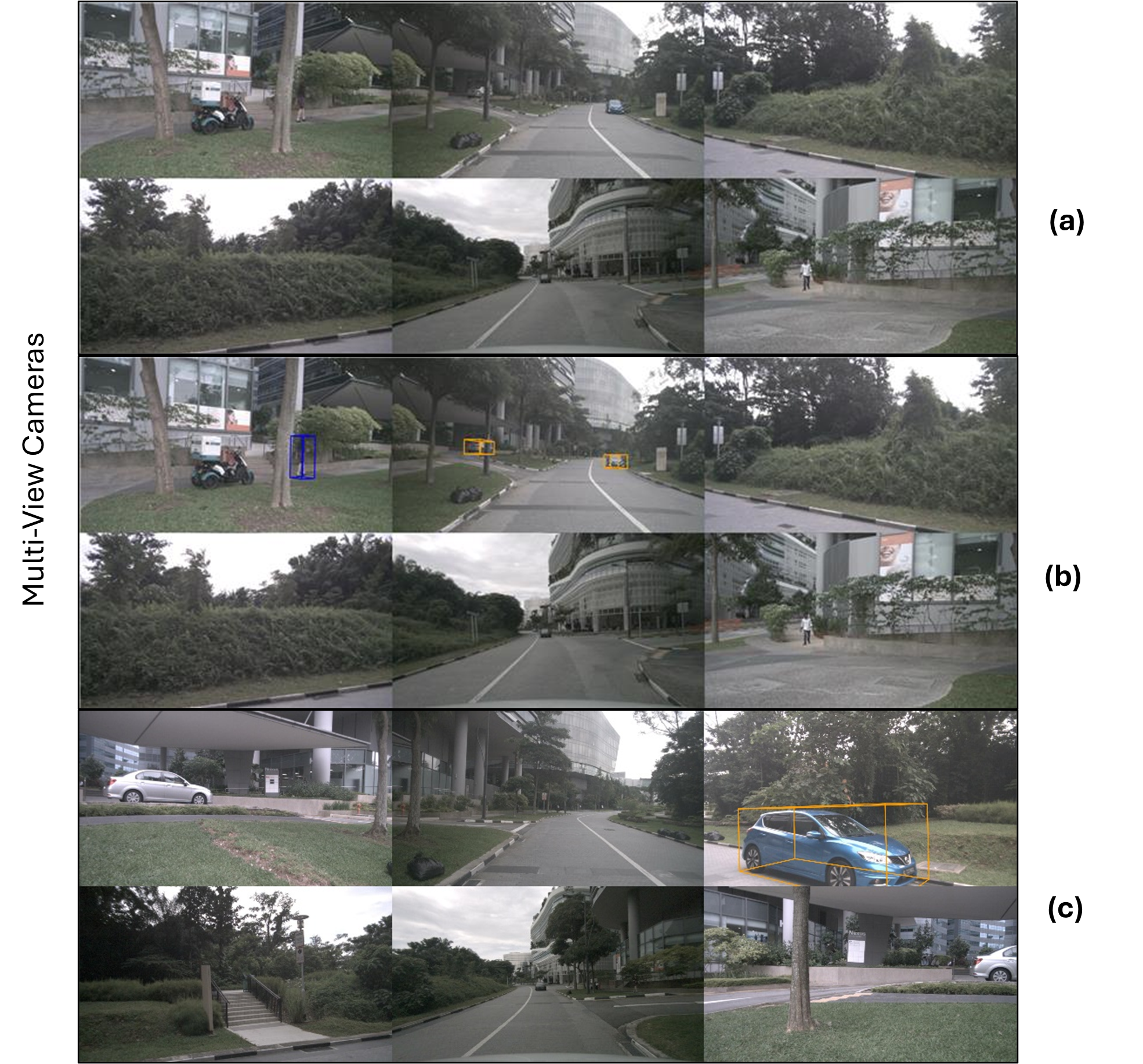}
    \vspace{-3mm}
    \caption{
    Qualitative comparison across nearby frames, which highlights a key observation regarding \gls{lidar} occlusion. Each row shows the six input camera images with overlaid 3D predictions: (a) \gls{lidar} only input, with heavy occlusion applied. (b) \gls{lidar} and camera input, with heavy occlusion applied to the \gls{lidar}. (c) \gls{lidar} only input, with the same heavy occlusion as (a), but applied to 5 samples later than previous rows. 
    }
\label{fig:lidar_imp_analysis}
\vspace{-5mm}
\end{figure}

Figure~\ref{fig:lidar_imp_analysis} illustrates the impact of \gls{lidar} occlusion on object detection across nearby frames. In (a), under heavy 90\% \gls{lidar} occlusion, the model fails to detect any objects in the current frame, especially those at greater distances. In (b), fusing the same occluded \gls{lidar} with clean camera input enables successful detection, highlighting the camera's contribution in recovering missed objects. In (c), using occluded \gls{lidar}-only again, the previously undetected object becomes visible and is detected after five frames, once it moves closer to the ego vehicle. This reveals a key insight: under severe occlusion, \gls{lidar} struggles with long-range detection but remains effective at short-range as objects approach, depending on the size and position of the occlusion. Overall, the figure demonstrates that camera input significantly enhances detection under \gls{lidar} degradation, while \gls{lidar} alone retains some utility for nearby objects.

\subsection{Quantitative Results}

In this section, we present the quantitative analysis of our study, focusing on 3D object detection under camera and \gls{lidar} occlusion, as shown in Table~\ref{tab:maintable}. The BEVFusion architecture is used as the baseline, with results reported for both unoccluded and occluded sensor configurations. When using only the camera (C), the baseline model achieves 35.6 mAP and 41.2 NDS. However, under camera occlusion, performance drops significantly to 20.9 mAP and 32.1 NDS, indicating the model’s sensitivity to visual degradation. For the \gls{lidar}-only setting (L), the baseline achieves significantly higher performance (64.7 mAP and 69.3 NDS), but under \gls{lidar} occlusion (L-occluded), these metrics sharply decline to 34.1 mAP and 49.6 NDS. This severe drop corresponds to a 90\% random dropout of \gls{lidar} point clouds, used to simulate extreme weather conditions such as dense fog or heavy rain. Since 3D object detection relies heavily on accurate depth information from \gls{lidar}, this drastic reduction in point density significantly impairs the model’s ability to localise and identify objects in 3D space.

In the fusion configuration (C+L), BEVFusion achieves the best performance with 68.5 mAP and 71.4 NDS. When only the camera is occluded in the fusion setup, the drop in performance is minimal, with mAP decreasing slightly to 65.7 and NDS to 70.0. This indicates that the model can effectively rely on \gls{lidar} data to compensate for missing visual information. In contrast, occluding the \gls{lidar} in the fusion configuration results in a more pronounced performance loss, with mAP falling to 50.1 and NDS to 57.9. These results highlight the critical role of \gls{lidar} in accurate 3D object detection and demonstrate that BEVFusion retains reasonable robustness under partial sensor occlusion, particularly when only the camera is degraded. \gls{lidar} is more sensitive to occlusion because it depends on dense geometric points for 3D localisation; when many points are lost, the spatial structure collapses. Cameras, even when partly occluded, usually retain enough semantic context to sustain detection.

To further analyse the effect of \gls{lidar} occlusion severity, we conducted experiments with 30\%, 60\%, 70\%, 80\%, and 90\% point cloud dropouts. As shown in Table~\ref{lidar_more_experiments}, performance in the \gls{lidar}-only setup remains relatively stable up to 60\% occlusion (mAP drops from 64.68 to 56.94). However, beyond 70\% occlusion, the drop becomes more significant, reaching 34.09 at 90\%. This indicates that \gls{lidar} remains fairly robust under moderate occlusion, but its reliability decreases sharply under severe degradation. In contrast, the sensor fusion setup (with a clean camera) maintains higher performance across all levels, with mAP dropping more gradually from 68.52 to 50.09. This demonstrates the benefit of multi-modal fusion, where clean camera input helps mitigate the impact of missing \gls{lidar} data. This improvement is due to BEVFusion’s ability to combine geometric information from \gls{lidar} with semantic context from camera images in a shared \gls{bev} space. Even under severe \gls{lidar} occlusion, the clean camera stream provides complementary cues that help maintain robust 3D detection.

\begin{table}[ht]
\vspace{-5mm}
\centering
\caption{Performance comparison of BEVFusion under different sensor occlusion settings. The table reports mAP\% and NDS\% for clean sensor inputs, as well as for scenarios with occluded camera, occluded \gls{lidar}, and both sensors combined with occlusion.}
\label{tab:maintable}
\begin{tabular}{lccc}
\toprule
\textbf{Method} & \textbf{Sensor} & \textbf{mAP\%} & \textbf{NDS\%} \\
\midrule
BEVFusion \cite{bib10} & C & 35.6 & 41.2  \\
\textbf{BEVFusion Occluded} & C-occluded & 20.9 & 32.1 \\
\midrule
BEVFusion \cite{bib10} & L & 64.7 & 69.3  \\
\textbf{BEVFusion Occluded } & L-occluded & 34.1 & 49.6 \\
\midrule
BEVFusion \cite{bib10} & C+L & 68.5 & 71.4  \\
\textbf{BEVFusion Occluded } & C+L (C-occluded) & 65.7 & 70.0 \\
\textbf{BEVFusion Occluded} & C+L (L-occluded) & 50.1 & 57.9 \\
\bottomrule
\end{tabular}
\vspace{-3mm}
\end{table}

\section{CONCLUSION AND FUTURE WORK}

In this work, we presented a systematic study of the impact of sensor occlusions on 3D object detection performance within a BEV-based multi-sensor fusion framework. Using the BEVFusion architecture, trained on clean nuScenes data, we evaluated detection accuracy under two types of occlusions: camera degradation, simulated by soiling masks to replicate lens obstructions, and \gls{lidar} degradation, achieved through random point cloud dropout to mimic visibility loss due to adverse weather conditions such as fog or rain. These occlusions were applied only during inference to isolate their individual and combined effects in a controlled setting. Our results show that camera-only models suffer significant performance drops even under moderate occlusion, while \gls{lidar}-only models remain relatively robust up to 60\% point dropout but degrade sharply beyond 70\%. The fusion setup demonstrates higher resilience across all conditions, with clean input from one modality helping to compensate for occlusion in the other. These findings underscore the importance of occlusion-aware evaluation and motivate the development of more robust fusion strategies that can maintain reliable performance under partial sensor failure or environmental degradation. In future work, we plan to apply sensor occlusions not only during testing but also during training, so the model can learn to handle degraded inputs more effectively. We also aim to explore the use of sequential frames (utilising temporal information) to help the model detect objects even when some sensors are momentarily occluded.

\section{ACKNOWLEDGMENTS}
This work was supported with the financial support of the Science Foundation Ireland grant 13/RC/2094\_P2 and co-funded under the European Regional Development Fund through the Southern \& Eastern Regional Operational Programme to Lero - the Science Foundation Ireland Research Centre for Software \href{www.lero.ie}{(www.lero.ie)}

\end{document}